# One-step and Two-step Classification for Abusive Language Detection on Twitter

Ji Ho Park and Pascale Fung
Human Language Technology Center
Department of Electronic and Computer Engineering
Hong Kong University of Science and Technology
jhpark@connect.ust.hk, pascale@ece.ust.hk

## Abstract

Automatic abusive language detection is a difficult but important task for online social media. Our research explores a two-step approach of performing classification on abusive language and then classifying into specific types and compares it with one-step approach of doing one multi-class classification for detecting sexist and racist languages. With a public English Twitter corpus of 20 thousand tweets in the type of sexism and racism, our approach shows a promising performance of 0.827 F-measure by using HybridCNN in one-step and 0.824 F-measure by using logistic regression in two-steps.

## 1 Introduction

Fighting abusive language online is becoming more and more important in a world where online social media plays a significant role in shaping people's minds (Perse and Lambe, 2016). Nevertheless, major social media companies like Twitter find it difficult to tackle this problem (Meyer, 2016), as the huge number of posts cannot be mediated with only human resources.

Warner and Hirschberg (2012) and Burnap and Williams (2015) are one of the early researches to use machine learning based classifiers for detecting abusive language. Djuric et al., (2015) incorporated representation word embeddings (Mikolov et al., 2013). Nobata et al. (2016) combined pre-defined language elements and word embedding to train a regression model. Waseem (2016) used logistic regression with n-grams and user-specific features such as gender and location. Davidson et al. (2017) conducted a deeper investigation on different types of abusive language. Badjatiya et al. (2017) experimented with deep learning-based models using ensemble gradient boost classifiers to perform multi-class classification on sexist and racist language. All approaches have been on one step.

Many have addressed the difficulty of the definition of abusive language while annotating the data, because they are often subjective to individuals (Ross et al. 2016) and lack of context (Waseem and Hovy, 2016; Schmidt & Wiegand, 2017). This makes it harder for non-experts to annotate without having a certain amount of domain knowledge (Waseem, 2016).

In this research, we aim to experiment a two-step approach of detecting abusive language first and then classifying into specific types and compare with a one-step approach of doing one multi-class classification on sexist and racist language.

Moreover, we explore applying a convolutional neural network (CNN) to tackle the task of abusive language detection. We use three kinds of CNN models that use both character-level and word-level inputs to perform classification on different dataset segmentations. We measure the performance and ability of each model to capture characteristics of abusive language.

## 2 Methodology

We propose to implement three CNN-based models to classify sexist and racist abusive language: CharCNN, WordCNN, and HybridCNN. The major difference among these models is whether the input features are characters, words, or both.

The key components are the convolutional layers that each computes a one-dimensional convolution over the previous input with multiple filter sizes and large feature map sizes. Having different filter sizes is the same as looking at a sentence with different windows simultaneously. Max-pooling is performed after the convolution to capture the feature that is most significant to the output.

## 2.1 CharCNN

CharCNN is a modification of the character-level convolutional network in (Zhang et al. 2015). Each character in the input sentence is first transformed into a one-hot encoding of 70 characters, including 26 English letters, 10 digits, 33 other characters, and a newline character (punctuations and special characters). All other non-standard characters are removed.

Zhang et al. (2015) uses 7 layers of convolutions and max-pooling layers, 2 fully-connected layers, and 1 softmax layer, but we also designed a shallow version with 2 convolutions and max-pooling layers, 1 fully-connected layers, and 1 softmax layers with dropout, due to the relatively small size of our dataset to prevent overfitting.

## 2.2 WordCNN

WordCNN is a CNN-static version proposed by Kim (2014). The input sentence is first segmented into words and converted into a 300-dimensional embedding *word2vec* trained on 100 billion words from Google News (Mikolov et al., 2013). Incorporating pre-trained vectors is a widely-used method to improve performance, especially when using a relatively small dataset. We set the embedding to be non-trainable since our dataset is small.

We propose to segment some out-of-vocabulary phrases as well. Since the Twitter tweets often contain hashtags such as *#womenagainstfeminism* and *#feminismisawful* we use a wordsegment library (Segaran and Hammerbacher, 2009) to capture more words.

## 2.3 HybridCNN

We design HybridCNN, a variation of WordCNN, since WordCNN has the limitation of only taking word features as input. Abusive language often contains either purposely or mistakenly misspelled words and made-up vocabularies such as *#feminazi*.

Therefore, since CharCNN and WordCNN do not use character and word inputs at the same time, we design the HybridCNN to experiment whether the model can capture features from both levels of inputs.

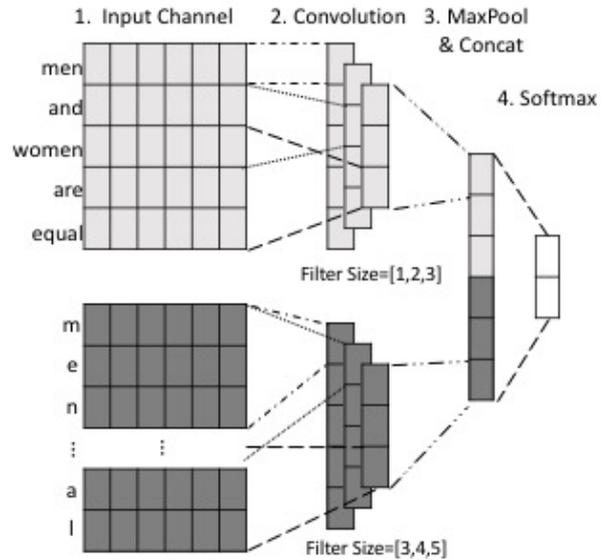

Figure 1 Architecture of HybridCNN

HybridCNN has two input channels. Each channel is fed into convolutional layers with three filter windows of different size. The output of the convolution are concatenated into one vector after 1-max-pooling. The vector is then fed into the final softmax layer to perform classification (See Figure 1).

## 3 Experiments

### 3.1 Datasets

We used the two English Twitter Datasets (Waseem and Hovy, 2016; Waseem, 2016) published as unshared tasks for the 1st Workshop on Abusive Language Online(ALW1). It contains tweets with sexist and racist comments. Waseem and Hovy (2016) created a list of criteria based on a critical race theory and let an expert annotate the corpus. First, we concatenated the two datasets into one and then divided that into three datasets for one-step and two-step classification (Table 1). One-step dataset is a segmentation for multi-class classification. For two-step classification, we merged the sexism and racism labels into one abusive label. Finally, we created another dataset with abusive languages to experiment a second classifier to distinguish "sexism" and "racism", given that the instance is classified as "abusive".

### 3.2 Training and Evaluation

We performed two classification experiments:
1. Detecting "none", "sexist", and "racist" language (one-step)

| Dataset | One-step | | | Two-step-1 | | Two-step-2 | |
|---|---|---|---|---|---|---|---|
| Label | None | Racism | Sexism | None | Abusive | Sexism | Racism |
| # | 12,427 | 2,059 | 3,864 | 12,427 | 5,923 | 2,059 | 3,864 |

Table 1: Dataset Segmentation

2. Detecting "abusive language", then further classifying into "sexist" or "racist" (two-step)

The purpose of these experiments was to see whether dividing the problem space into two steps makes the detection more effective.

We trained the models using mini-batch stochastic gradient descent with AdamOptimizer (Kingma and Ba, 2014). For more efficient training in an unbalanced dataset, the mini-batch with a size of 32 had been sampled with equal distribution for all labels. The training continued until the evaluation set loss did not decrease any longer. All the results are average results of 10-fold cross validation.

As evaluation metric, we used F1 scores with precision and recall score and weighted averaged the scores to consider the imbalance of the labels. For this reason, total average F1 might not between average precision and recall.

As baseline, we used the character n-gram logistic regression classifier (indicated as LR on Table 2-4) from Waseem and Hovy (2016), Support Vector Machines (SVM) classifier, and FastText (Joulin et al., 2016) that uses average bag-of-words representations to classify sentences. It was the second best single model on the same dataset after CNN (Badjatiya et al., 2017).

### 3.3 Hyperparameters

For hyperparameter tuning, we evaluated on the validation set. These are the hyperparmeters used for evaluation.
- **CharCNN:** Shallow model with 1024 feature units for convolution layers with filter size 4, max-pooling size 3, and L2 regularization constant 1 and 2048 units for the fully-connected layer
- **WordCNN:** Convolution layers with 3 filters with the size of [1,2,3] and feature map size 50, max-pooling, and L2 regularization constant 1
- **HybridCNN:** For the character input channel, convolution layers with 3 filters with size of [3,4,5] and for word input channel, 3 filters with size of [1,2,3]. Both channels had feature map size of 50, max-pooling, and L2 regularization constant 1.

## 4 Result and Discussions

### 4.1 One-step Classification

The results of the one-step multi-class classification are shown in the top part of Table 2.

Our newly proposed HybridCNN performs the best, giving an improvement over the result from WordCNN. We expected the additional character input channel improves the performance. We assumed that the reason CharCNN performing worse than WordCNN is that the dataset is too small for the character-based model to capture word-level features by itself.

Baseline methods tend to have high averaged F1 but low scores on racism and sexism labels due to low recall scores.

### 4.2 Two-step Classification

The two-step approach that combines two binary classifiers shows comparable results with one-step approach. The results of combining the two are shown in the bottom part of Table 3.

Combining two logistic regression classifiers in the two-step approach performs about as well as one-step HybridCNN and outperform one-step logistic regression classifier by more than 10 F1 points. This is surprising since logistic regression takes less features than the HybridCNN.

Furthermore, using HybridCNN on the first step to detect abusive language and logistic regression on the second step to classify racism and sexism worked better than just using HybridCNN.

Table 4 shows the results of abusive language classification. HybridCNN also performs best for abusive language detection, followed by WordCNN and logistic regression.

Table 5 shows the results of classifying into sexism and racism given that it is abusive. The second classifier has significant performance in predicting a specific type (in this case, sexism and racism) of an abusive language. We can deduce that sexist and racist comments have obvious discriminating features that are easy for all classifiers to capture.

| | None | | | Racism | | | Sexism | | | Total | | |
|---|---|---|---|---|---|---|---|---|---|---|---|---|
| Method | Prec. | Rec. | F1 | Prec. | Rec. | F1 | Prec. | Rec. | F1 | Prec. | Rec. | F1 |
| LR | .824 | .945 | .881 | .810 | .598 | .687 | .835 | .556 | .668 | .825 | .824 | .814 |
| SVM | .802 | **.956** | .872 | **.815** | .531 | .643 | **.851** | .483 | .616 | .814 | .808 | .793 |
| FastText | .828 | .922 | **.882** | .759 | .630 | .685 | .777 | .557 | .648 | .810 | .812 | .804 |
| CharCNN | .861 | .867 | .864 | .693 | .746 | .718 | .713 | .666 | .688 | .801 | .811 | .811 |
| WordCNN | .870 | .868 | .868 | .704 | .762 | .731 | .712 | **.686** | .694 | .818 | .816 | .816 |
| HybridCNN | **.872** | .882 | .877 | .713 | **.766** | **.736** | .743 | .679 | **.709** | **.827** | **.827** | **.827** |
| LR (two) | .841 | .933 | **.895** | .800 | .664 | .731 | .809 | .590 | .683 | **.828** | **.831** | .824 |
| SVM (two) | .816 | **.945** | .876 | **.811** | .605 | .689 | **.823** | .511 | .630 | .816 | .815 | .803 |
| HybridCNN (two) | .877 | .864 | .869 | .690 | **.759** | .721 | .705 | .701 | **.699** | .807 | .809 | .807 |
| HybridCNN + LR(two) | **.880** | .859 | .869 | .722 | .751 | **.735** | .683 | **.717** | **.699** | .821 | .817 | .818 |

Table 2. Experiment Results: upper part is the one-step methods that perform multi-class classification and lower methods with (two) indicate two-step that combines two binary classifiers. HybridCNN is our newly created model.

Since the precision and recall scores of the "abusive" label is higher than those of "racism" and "sexism" in the one-step approach, the two-step approach can perform as well as the one-step approach.

| **Model** | Prec. | Rec. | F1 |
|---|---|---|---|
| LR | .816 | .640 | .711 |
| SVM | **.839** | .560 | .668 |
| FastText | .765 | .616 | .683 |
| CharCNN | .743 | .674 | .707 |
| WordCNN | .731 | .722 | .726 |
| HybridCNN | .719 | **.754** | **.734** |

Table 3. Results on Abusive Language Detection

| **Model** | Prec. | Rec. | F1 |
|---|---|---|---|
| LR | **.954** | **.953** | **.952** |
| SVM | .954 | .953 | .952 |
| FastText | .937 | .937 | .937 |
| CharCNN | .941 | .941 | .941 |
| WordCNN | .952 | .952 | .952 |
| HybridCNN | .951 | .950 | .950 |

Table 4. Results on Sexist/Racist Classification

## 5 Conclusion and Future work

We explored a two-step approach of combining two classifiers - one to classify abusive language and another to classify a specific type of sexist and racist comments given that the language is abusive. With many different machine learning classifiers including our proposed HybridCNN, which takes both character and word features as input, we showed the potential in the two-step approach compared to the one-step approach which is simply a multi-class classification. In this way, we can boost the performance of simpler models like logistic regression, which is faster and easier to train, and combine different types of classifiers like convolutional neural network and logistic regression together depending on each of its performance on different datasets.

We believe that two-step approach has potential in that large abusive language datasets with specific label such as profanity, sexist, racist, homophobic, etc. is more difficult to acquire than those simply flagged as abusive.

For this reason, in the future we would like to explore training the two-step classifiers on separate datasets (for example, a large dataset with abusive language for the first-step classifier and smaller specific-labelled dataset for the second-step classifier) to build a more robust and detailed abusive language detector.

## Acknowledgements

This work is partially funded by CERG16214415 of the Hong Kong Research Grants Council and ITS170 of the Innovation and Technology Commission.

## References

Badjatiya, P., Gupta, S., Gupta, M., & Varma, V. (2017). Deep learning for hate speech detection in tweets. *Proceedings of the 26th International Conference on World Wide Web Companion,* 759-760.

Djuric, N., Zhou, J., Morris, R., Grbovic, M., Radosavljevic, V., & Bhamidipati, N. (2015). Hate speech detection with comment embeddings. *Proceedings of the 24th International Conference on World Wide Web,* 29-30.

Joulin, A., Grave, E., Bojanowski, P., & Mikolov, T. (2016). Bag of tricks for efficient text classification. *Proceedings of the 15th Conference of the European Chapter of the Association for Computational Linguistics: Volume 2, Short Papers,*

Kim, Y. (2014). Convolutional neural networks for sentence classification . *In Proceedings of EMNLP.,*

Kingma, D., & Ba, J. (2014). Adam: A method for stochastic optimization. *Proceedings of the 3rd International Conference on Learning Representations (ICLR)*

LeCun, Y., Kavukcuoglu, K., & Farabet, C. (2010). Convolutional networks and applications in vision. *Circuits and Systems (ISCAS), Proceedings of 2010 IEEE International Symposium on,* 253-256.

Meyer, R. (2016, 07/21). Twitter's famous racist problem. *The Atlantic,*

Mikolov, T., Sutskever, I., Chen, K., Corrado, G. S., & Dean, J. (2013). Distributed representations of words and phrases and their compositionality. *Advances in Neural Information Processing Systems,* 3111-3119.

Nobata, C., Tetreault, J., Thomas, A., Mehdad, Y., & Chang, Y. (2016). Abusive language detection in online user content. *Proceedings of the 25th International Conference on World Wide Web,* 145-153.

Perse, E. M., & Lambe, J. (2016). *Media effects and society* Routledge.

Ross, B., Rist, M., Carbonell, G., Cabrera, B., Kurowsky, N., & Wojatzki, M. (2017). Measuring the reliability of hate speech annotations: The case of the european refugee crisis In *Proceedings of the Workshop on Natural Language Processing for ComputerMediated Communication (NLP4CMC), pages 6–9*

Schmidt, A., & Wiegand, M. (2017). A survey on hate speech detection using natural language processing. *Socialnlp 2017, ,* 1.

Segaran, T., & Hammerbacher, J. (2009). *Beautiful data: The stories behind elegant data solutions* " O'Reilly Media, Inc.".

Wang, W., Chen, L., Thirunarayan, K., & Sheth, A. P. (2012). Harnessing twitter" big data" for automatic emotion identification. *Privacy, Security, Risk and Trust (PASSAT), 2012 International Conference on and 2012 International Confernece on Social Computing (SocialCom),* 587-592.

Warner, W., & Hirschberg, J. (2012). Detecting hate speech on the world wide web. *Proceedings of the Second Workshop on Language in Social Media,* 19-26.

Waseem, Z. (2016). Are you a racist or am I seeing things? annotator influence on hate speech detection on twitter. *Proceedings of the 1st Workshop on Natural Language Processing and Computational Social Science,* 138-142.

Waseem, Z., & Hovy, D. (2016). Hateful symbols or hateful people? predictive features for hate speech detection on twitter. *Proceedings of NAACL-HLT,* 88-93.

Zhang, X., Zhao, J., & LeCun, Y. (2015). Character-level convolutional networks for text classification. *Advances in Neural Information Processing Systems,* 649-657.